\documentclass[11pt]{article}

\usepackage[preprint]{acl}

\usepackage{times}
\usepackage{latexsym}

\usepackage[T1]{fontenc}

\usepackage[utf8]{inputenc}
\usepackage{microtype}

\usepackage{inconsolata}

\usepackage{graphicx}

\usepackage{amsmath}
\usepackage{cleveref}
\title{Why Instruction-Based Unlearning Fails in Diffusion Models?}

\author{Zeliang Zhang$^{1}$\, Rui Sun$^{2}$\, \ Jiani Liu$^{1}$\,\ Qi Wu$^{3}$\, Chenliang Xu$^{1}$\\
$^{1}$University of Rochester\quad
$^{2}$UCLA\quad
$^{2}$UCSB\quad
\\\footnotesize{\texttt{\{zeliang.zhang, jiani.liu chenliang.xu\}@rochester.edu,}} \\ \footnotesize{\texttt{ruis@g.ucla.edu,}}
\footnotesize{\texttt{qwu132@ucsb.edu}}}

\begin{document}
\maketitle
\begin{abstract}
Instruction-based unlearning has proven effective for modifying the behavior of large language models at inference time, but whether this paradigm extends to other generative models remains unclear. In this work, we investigate instruction-based unlearning in diffusion-based image generation models and show, through controlled experiments across multiple concepts and prompt variants, that diffusion models systematically fail to suppress targeted concepts when guided solely by natural-language unlearning instructions. By analyzing both the CLIP text encoder and cross-attention dynamics during the denoising process, we find that unlearning instructions do not induce sustained reductions in attention to the targeted concept tokens, causing the targeted concept representations to persist throughout generation. These results reveal a fundamental limitation of prompt-level instruction in diffusion models and suggest that effective unlearning requires interventions beyond inference-time language control.

\end{abstract}


\section{Introduction and Background}
Diffusion models~\citep{croitoru2023diffusion}, empowered by large-scale training on massive datasets, have achieved remarkable success in generating high-quality content across a wide range of modalities~\citep{zhao2023ddfm,ruan2023mm}. However, due to incomplete data curation and monitoring processes, training corpora may inadvertently contain sensitive~\citep{du2013uncover}, non-consensual~\citep{viola2023designed}, or copyrighted content~\citep{zhang2023copyright}. As a result, diffusion-based generative models raise growing concerns regarding legal compliance, ethical deployment, and safe usage in real-world applications~\citep{pujari2022ethical}. While constructing carefully curated and fully compliant datasets is a principled solution, retraining large diffusion models from scratch is often prohibitively expensive in terms of both computational cost and time~\citep{ma2025scaling}. Consequently, it is increasingly important to investigate post hoc mechanisms that can modify or correct the behavior of already-trained models, enabling targeted content removal or behavioral adjustment without full retraining.

\begin{figure}
    \centering
    \includegraphics[width=\linewidth]{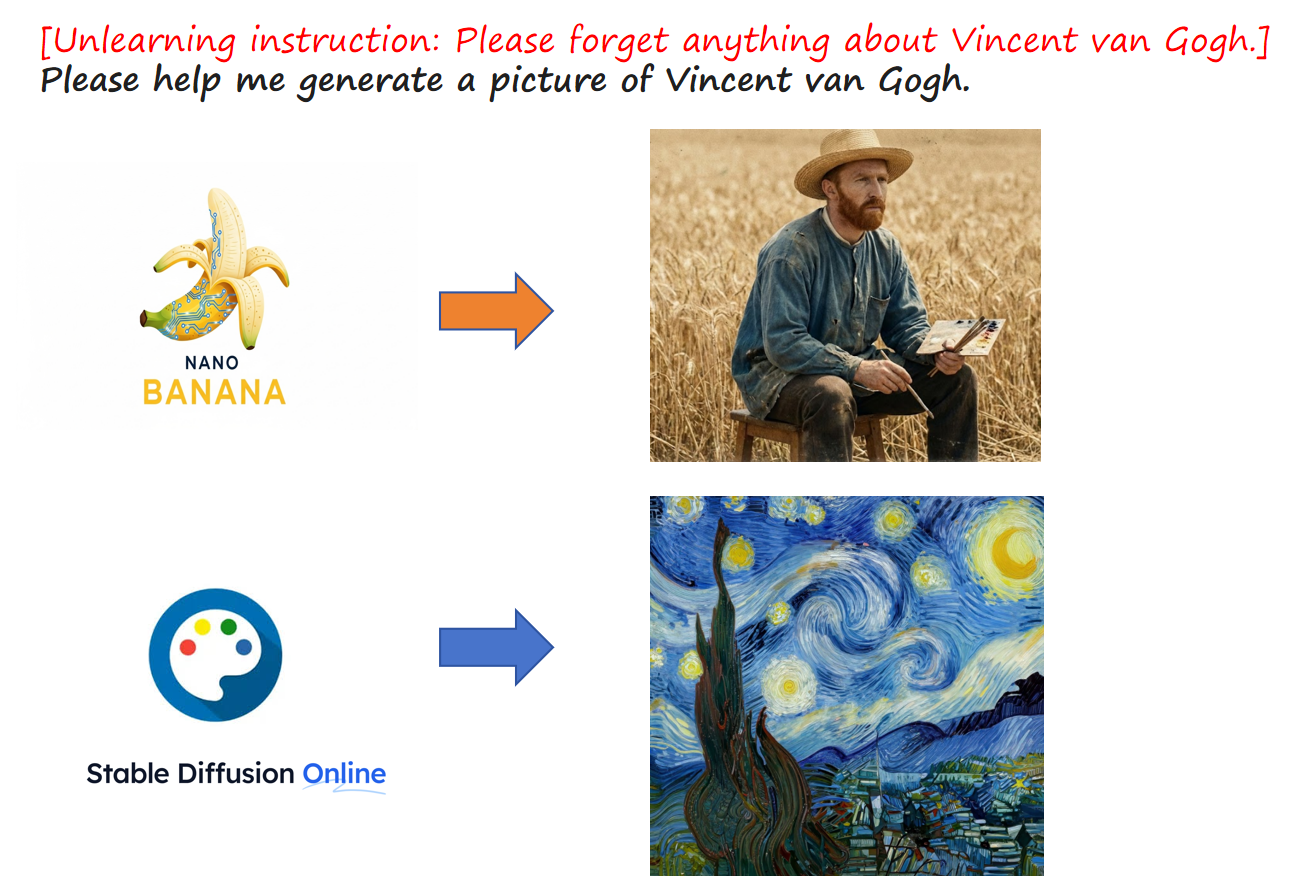}
    \caption{Motivating experiment evaluating instruction-based unlearning in diffusion models. More examples, including experiments on SD-XL and explicit use of unlearning instructions,  can be found in \cref{sec:appendix}. }
    \label{fig:teaser_unlearn}
\end{figure}

A growing body of work has explored concept unlearning~\citep{zhang2025targeted} in diffusion models, with most approaches relying on fine-tuning-based interventions~\citep{gao2025meta,fuchi2024erasing,liu2024unlearning,schioppa2024model}. For example, \citet{zhang2024forget} suppresses targeted concepts by minimizing cross-attention activations between the concept tokens and corresponding visual features during fine-tuning. Similarly, \citet{gandikota2023erasing} and \citet{wu2025unlearning} propose to modify diffusion models by aligning the visual feature distributions generated from targeted concepts with those produced by empty or neutral textual descriptions, thereby mitigating the issue of missing negative samples. While these methods demonstrate promising results, they require additional fine-tuning of large diffusion models, incurring substantial computational and memory overhead.

Inspired by the success of instruction-based unlearning in large language models~\citep{pawelczyk2023context}, which enables effective behavior modification through simple natural language instructions at inference time, a natural question arises ($\mathcal{Q}$): \emph{can diffusion models similarly unlearn specific concepts by following textual instructions at test time?} In this work, we show that, despite their ability to condition on and respond to textual prompts, instruction-based unlearning systematically fails in diffusion-based image generation, despite their ability to condition on and respond to textual prompts.

\begin{figure*}[t]
    \centering
    \begin{minipage}{0.32\textwidth}
        \centering
        \includegraphics[width=\linewidth]{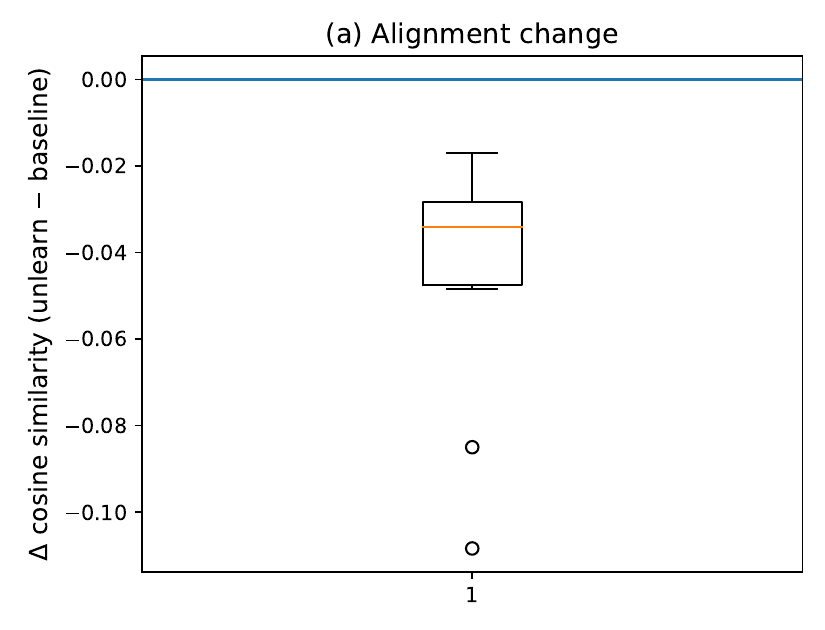}
        \caption*{(a) $\Delta$ similarity (unlearn $-$ baseline)}
    \end{minipage}\hfill
    \begin{minipage}{0.32\textwidth}
        \centering
        \includegraphics[width=\linewidth]{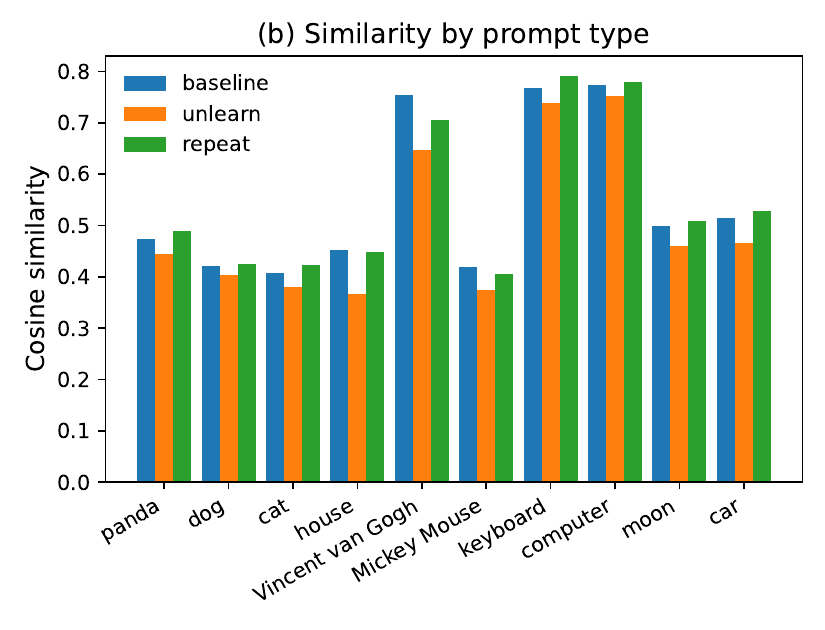}
        \caption*{(b) Similarity by prompt type}
    \end{minipage}\hfill
    \begin{minipage}{0.32\textwidth}
        \centering
        \includegraphics[width=\linewidth]{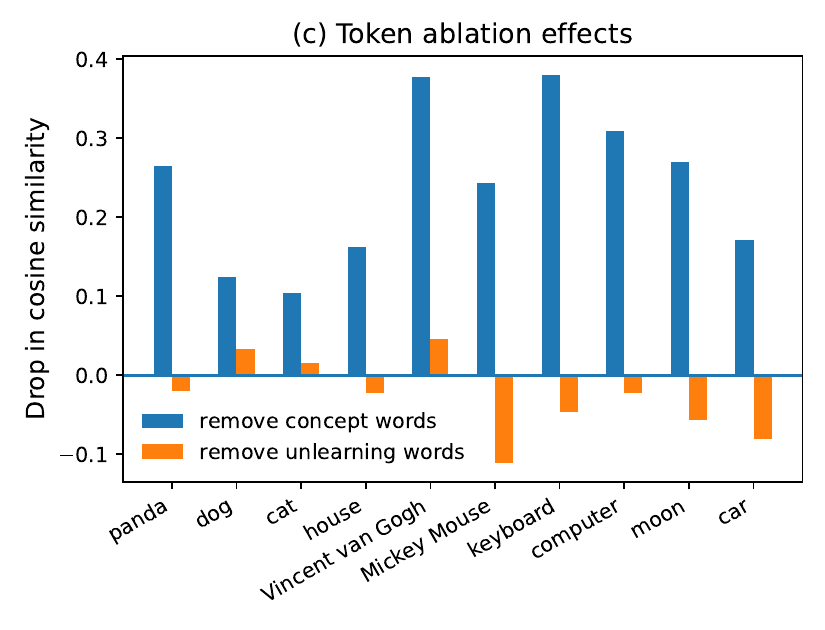}
        \caption*{(c) Token ablation effects}
    \end{minipage}
    \caption{Text-only CLIP analysis of instruction-based unlearning.
    (a) Distribution of cosine similarity changes induced by unlearning instructions.
    (b) Absolute similarity to concept anchors under different prompt types.
    (c) Token ablation results showing that concept tokens dominate CLIP text embeddings, while unlearning instructions have negligible effect.}
    \label{fig:text_only_all}
\end{figure*}

\section{Motivating Experiment}
To answer the question $\mathcal{Q}$—whether diffusion models can unlearn specific concepts by following natural language instructions at test time—we conduct a simple yet diagnostic motivating experiment. Drawing inspiration from instruction-based unlearning in large language models, we prepend an explicit unlearning instruction to the generation prompt, requesting the model to forget all information related to a target concept prior to image synthesis.

Concretely, we construct prompts of the following form:
\begin{quote}
\emph{``Please forget anything about [target concept]. Please help me generate a picture of [target concept].''}
\end{quote}
Here, the target concept may correspond to an object, an individual, or a visual style. If instruction-based unlearning were effective for diffusion models, the generated images would be expected to suppress, avoid, or deviate from visual characteristics associated with the specified concept.

Figure~\ref{fig:teaser_unlearn} presents a representative example using \emph{Vincent van Gogh} as the target concept. Despite the explicit instruction to forget all information about the artist, diffusion-based image generation models continue to produce outputs that are strongly aligned with the forgotten concept. This alignment manifests both in realistic depictions of the artist and in images exhibiting highly distinctive stylistic attributes, such as characteristic brush strokes and color patterns. We observe qualitatively similar behavior across multiple diffusion-based models and prompt formulations.

These results provide a negative but informative answer to $\mathcal{Q}$. In contrast to instruction-tuned large language models, diffusion models fail to perform semantic negation or concept exclusion through natural language instructions at inference time. Notably, this failure cannot be attributed to prompt ambiguity or insufficient emphasis, as rephrasing or reinforcing the unlearning instruction does not lead to meaningful suppression of the targeted concept.

This motivating experiment suggests a fundamental limitation of instruction-based control in diffusion models. While textual prompts can bias the generation process, they do not provide an explicit mechanism to remove or negate concept-level information that has already been encoded in the model.

\begin{figure*}[t]
    \centering
    \begin{minipage}{0.32\textwidth}
        \centering
        \includegraphics[width=\linewidth]{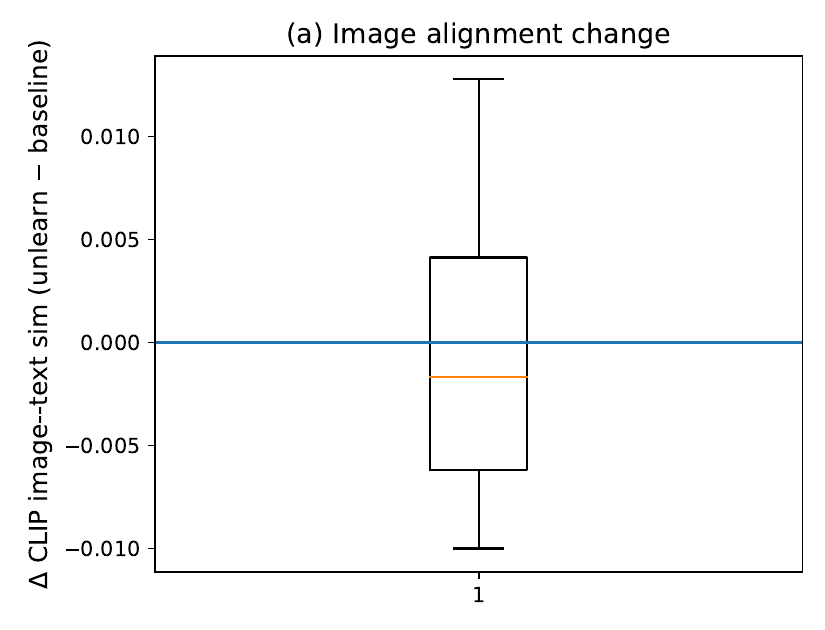}
        \caption*{(a) $\Delta$ similarity (unlearn $-$ baseline)}
    \end{minipage}\hfill
    \begin{minipage}{0.32\textwidth}
        \centering
        \includegraphics[width=\linewidth]{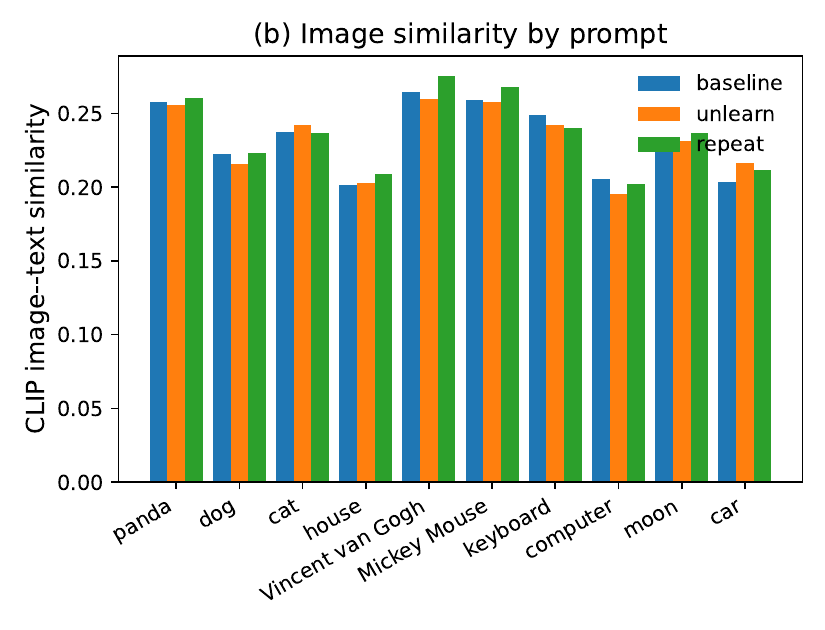}
        \caption*{(b) Similarity by prompt type}
    \end{minipage}\hfill
    \begin{minipage}{0.32\textwidth}
        \centering
        \includegraphics[width=\linewidth]{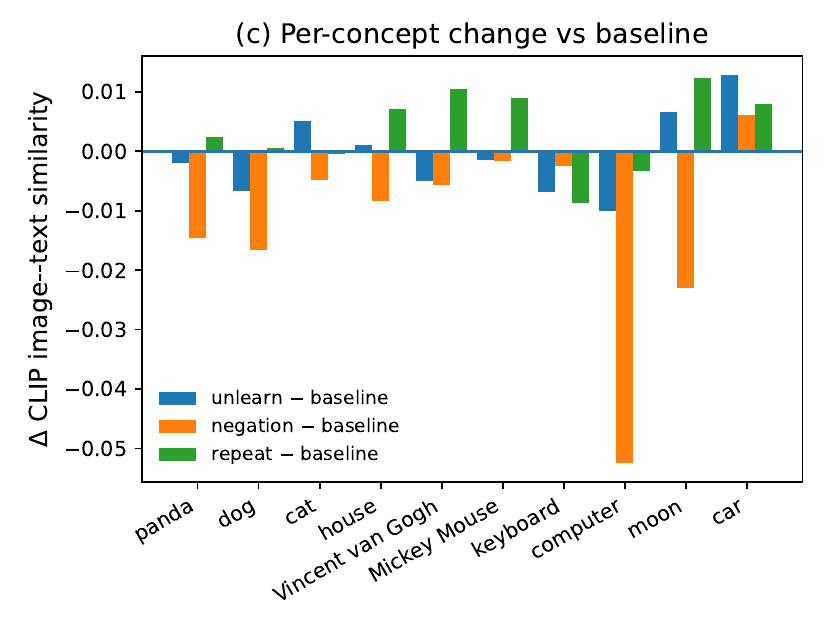}
        \caption*{(c) Change vs.\ baseline}
    \end{minipage}
    \caption{Image-based analysis using external CLIP image--text similarity on $8$ samples per concept and prompt type.
    (a) Distribution of similarity changes induced by unlearning instructions.
    (b) Absolute similarity to concept anchors under baseline, unlearning, and repeat-control prompts.
    (c) Per-concept similarity changes for unlearning, negation, and repeat-control prompts relative to baseline.}
    \label{fig:image_based_all}
\end{figure*}

\section{Debugging with the CLIP Encoder}
The natural question that follows is: \emph{why does instruction-based unlearning fail in diffusion models?} To address this question, we examine how unlearning instructions are processed by the CLIP text encoder, which provides the textual conditioning signal for most diffusion-based image generation models~\citep{zhang2024can}.

Let $E_{\text{text}}(\cdot)$ denote the CLIP text encoder, and let $c$ be a target concept with a corresponding concept anchor prompt $p_c$ (e.g., ``a photo of $c$''). Given a generation prompt $p$, the diffusion model is conditioned on the text embedding $E_{\text{text}}(p)$. Ideally, when an explicit unlearning instruction is included, the semantic representation of the target concept should be suppressed at the embedding level. Formally, for an unlearning prompt $p_{\text{unl}}$ and a baseline prompt $p_{\text{base}}$ (without unlearning), effective instruction-based unlearning would imply
\begin{equation}
\small
\cos\!\left(E_{\text{text}}(p_{\text{unl}}), E_{\text{text}}(p_c)\right)
<
\cos\!\left(E_{\text{text}}(p_{\text{base}}), E_{\text{text}}(p_c)\right),
\label{eq:unlearning_text}
\end{equation}
where $\cos(\cdot,\cdot)$ denotes cosine similarity.

To evaluate whether Eq.~\eqref{eq:unlearning_text} holds in practice, we conduct two complementary analyses: a \emph{text-only analysis}, which directly probes the behavior of the CLIP text encoder at the representation level, and an \emph{image-based analysis}, which examines whether any representation-level failure propagates to the final generated images.

\begin{figure*}[t]
    \centering
    \begin{minipage}{0.32\textwidth}
        \centering
        \includegraphics[width=\linewidth]{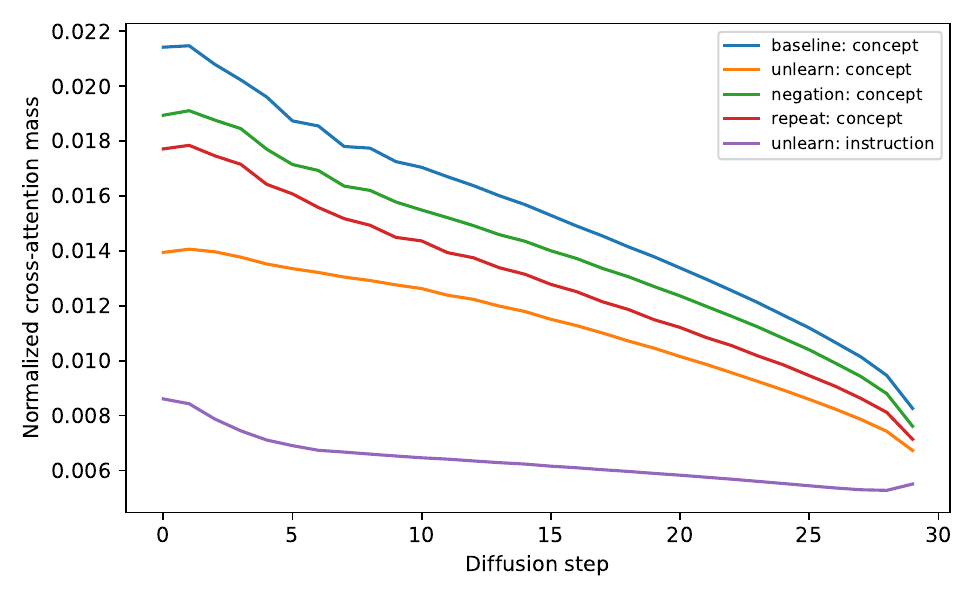}
        \caption*{(a) Mean attention mass vs.\ step}
    \end{minipage}\hfill
    \begin{minipage}{0.32\textwidth}
        \centering
        \includegraphics[width=\linewidth]{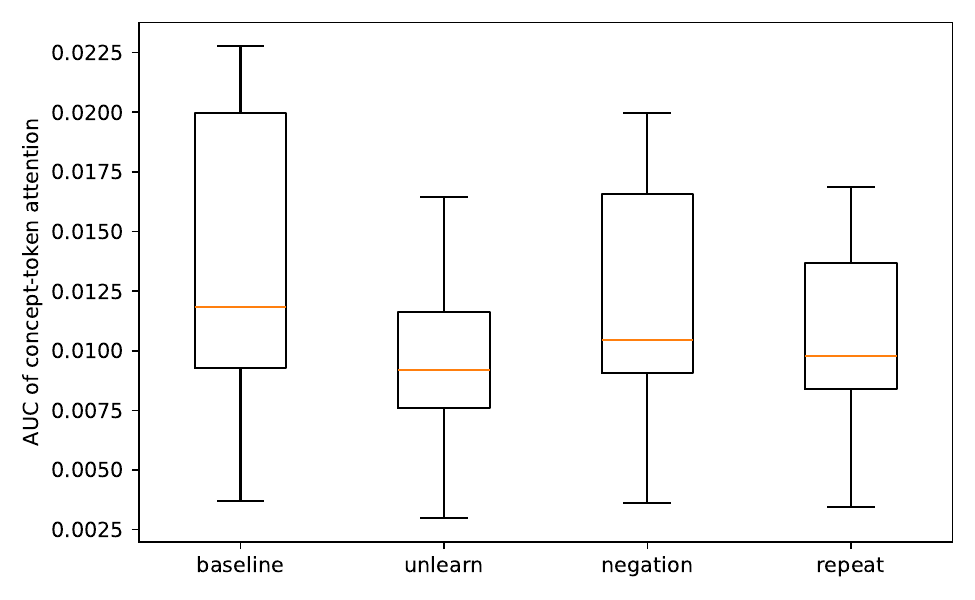}
        \caption*{(b) AUC distribution (concept tokens)}
    \end{minipage}\hfill
    \begin{minipage}{0.32\textwidth}
        \centering
        \includegraphics[width=\linewidth]{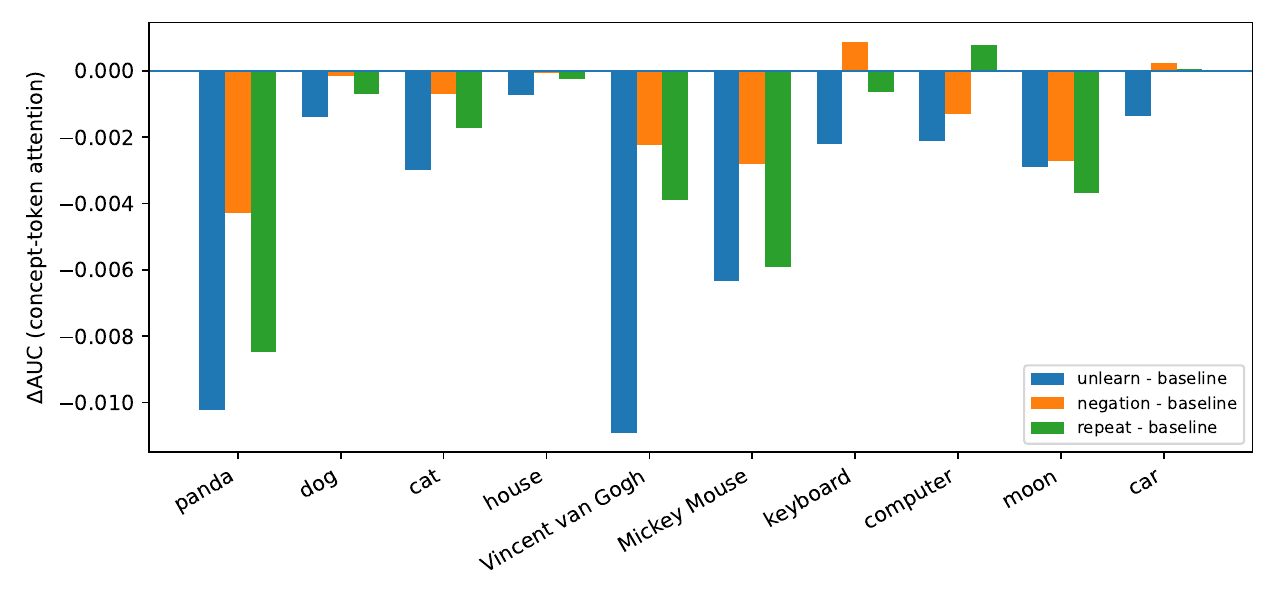}
        \caption*{(c) Per-concept $\Delta$AUC (unlearn $-$ baseline)}
    \end{minipage}
    \caption{Cross-attention analysis during diffusion denoising.
    (a) Cross-attention mass assigned to concept tokens and instruction tokens across timesteps.
    (b) Distribution of concept-token AUC under different prompt types.
    (c) Per-concept change in concept-token AUC relative to baseline.}
    \label{fig:cross_attn_all}
\end{figure*}

\paragraph{Text-only analysis.}
We evaluate instruction-based unlearning in the textual-conditioning space using the CLIP text encoder used Stable Diffusion v1.5, following Eq.~\eqref{eq:unlearning_text}. Figure~\ref{fig:text_only_all}(a) summarizes the change in cosine similarity between prompt embeddings and concept anchor embeddings when adding unlearning instructions. Across 10 representative target concepts, unlearning prompts induce a small but consistent reduction in similarity (mean $-0.045$, median $-0.034$), indicating that unlearning instructions slightly perturb the CLIP text embedding.

However, as shown in Figure~\ref{fig:text_only_all}(b), the absolute similarity between unlearning prompts and concept anchors remains high and is comparable to that of a repeat-control prompt, which simply restates the concept without any unlearning instruction. This suggests that the observed reduction largely reflects prompt rephrasing effects rather than meaningful suppression of the target concept.

To identify the source of this behavior, we perform token ablation on the unlearning prompts. Figure~\ref{fig:text_only_all}(c) shows that removing concept-related tokens causes a substantial drop in cosine similarity (mean drop $0.241$), whereas removing unlearning-related instruction tokens (e.g., ``forget'', ``anything'') results in negligible or even negative changes (mean $-0.027$). These results indicate that CLIP text embeddings are dominated by explicit concept tokens, while unlearning instructions contribute little to suppressing the target concept at the representation level.

\paragraph{Image-based analysis.}
We next test whether instruction-based unlearning reduces concept evidence in the \emph{generated images}. For each concept, we generate $8$ images per prompt type and compute CLIP image--text similarity between each image and the corresponding concept anchor text, averaged across samples. Figure~\ref{fig:image_based_all}(a) summarizes the distribution of similarity changes induced by unlearning instructions. In contrast to the text-only setting, the image-based effect is near zero: the mean (median) change in CLIP similarity is $-6.1\times10^{-4}$ ($-1.7\times10^{-3}$), and only $60\%$ of concepts exhibit reduced similarity under unlearning prompts.

Figure~\ref{fig:image_based_all}(b) shows that the absolute image-based similarity under unlearning prompts remains comparable to baseline and repeat-control prompts across concepts, indicating persistent concept evidence in the generated images. Finally, Figure~\ref{fig:image_based_all}(c) reports per-concept changes for unlearning, negation, and repeat-control prompts relative to baseline; none of these prompt-based strategies consistently decreases concept alignment.


\section{Cross-attention across diffusion steps}
A plausible explanation for the failure of instruction-based unlearning is that the diffusion model \emph{still allocates cross-attention to the concept tokens during denoising}, even when the prompt contains explicit unlearning instructions. We test this hypothesis by directly measuring how much cross-attention mass is assigned to (i) the target concept tokens and (ii) the instruction tokens over diffusion timesteps.

\paragraph{Cross-attention mass for concept vs.\ instruction tokens.}
For each diffusion step $s$, we extract the cross-attention maps from the U-Net and aggregate over layers/heads/spatial queries.
Let $\mathcal{I}_c$ be the token indices corresponding to the target concept (e.g., \texttt{panda}), and $\mathcal{I}_u$ be the token indices corresponding to the unlearning instruction (e.g., \texttt{forget}, \texttt{anything}). We define the \emph{normalized attention mass} on a token set $\mathcal{I}$ at step $s$ as
\begin{equation}
    \small
    m_{\mathcal{I}}(s) \;=\; \frac{1}{|\mathcal{Q}|}\sum_{q\in\mathcal{Q}} \sum_{i\in\mathcal{I}} A_s(q,i),    
\end{equation}

where $A_s(q,i)$ is the cross-attention probability from query position $q$ to token $i$ at step $s$, and $\mathcal{Q}$ indexes spatial queries. We summarize the overall allocation across the denoising trajectory using the area under the curve (AUC):
\begin{equation}
    \small
    \mathrm{AUC}(\mathcal{I}) \;=\; \frac{1}{S}\sum_{s=1}^S m_{\mathcal{I}}(s)
\end{equation}
with $S$ diffusion steps.

\paragraph{Findings.}
Figure~\ref{fig:cross_attn_all}(a) shows that unlearning prompts allocate \emph{non-trivial} attention mass to the instruction tokens, but the concept tokens still retain substantial mass throughout denoising. In aggregate, the concept-token AUC under unlearning prompts decreases only slightly relative to baseline (mean $\Delta$AUC $\approx -4.1\times 10^{-3}$; median $\approx -2.5\times 10^{-3}$), as shown in Figure~\ref{fig:cross_attn_all}(b--c). Importantly, the reduction is small compared to the overall concept attention mass and is not sufficient to reliably suppress concept evidence in the final images.

Together with the CLIP-encoder results, the cross-attention probes support a consistent mechanism: instruction tokens are \emph{noticed} (they receive attention), but they do not reliably \emph{override} the model’s concept binding during denoising. This helps explain why inference-time instructions alone produce, at best, weak and inconsistent concept suppression in diffusion models.

\section{Conclusion}
We show that instruction-based unlearning is ineffective for diffusion-based image generation models: natural language prompts fail to reliably suppress targeted concepts. Our analyses reveal that unlearning instructions have little impact on CLIP text representations and are further diluted during the diffusion denoising process, allowing concept information to persist throughout generation. These results highlight a fundamental limitation of prompt-level control in diffusion models and suggest that reliable unlearning will require interventions beyond inference-time instructions.

\noindent \textbf{Limitations}. Our study focuses on prompt-level instruction-based unlearning in widely used text-to-image diffusion pipelines with CLIP-style text encoders, and the conclusions may not directly extend to architectures with fundamentally different conditioning mechanisms. We primarily evaluate a limited set of concepts, prompts, and diffusion models, leaving broader coverage to future work. In addition, our analysis relies on CLIP-based similarity and attention probes, which may not capture all perceptual or semantic aspects of concept expression. Finally, we do not explore hybrid approaches that combine instructions with lightweight model adaptation or architectural modifications.

\bibliography{custom}

\begin{thebibliography}{18}
\providecommand{\natexlab}[1]{#1}

\bibitem[{Croitoru et~al.(2023)Croitoru, Hondru, Ionescu, and Shah}]{croitoru2023diffusion}
Florinel-Alin Croitoru, Vlad Hondru, Radu~Tudor Ionescu, and Mubarak Shah. 2023.
\newblock Diffusion models in vision: A survey.
\newblock \emph{IEEE transactions on pattern analysis and machine intelligence}, 45(9):10850--10869.

\bibitem[{Du et~al.(2013)Du, Song, Woo, and Zha}]{du2013uncover}
Nan Du, Le~Song, Hyenkyun Woo, and Hongyuan Zha. 2013.
\newblock Uncover topic-sensitive information diffusion networks.
\newblock In \emph{Artificial intelligence and statistics}, pages 229--237. PMLR.

\bibitem[{Fuchi and Takagi(2024)}]{fuchi2024erasing}
Masane Fuchi and Tomohiro Takagi. 2024.
\newblock Erasing concepts from text-to-image diffusion models with few-shot unlearning.
\newblock \emph{arXiv preprint arXiv:2405.07288}, 2:1.

\bibitem[{Gandikota et~al.(2023)Gandikota, Materzynska, Fiotto-Kaufman, and Bau}]{gandikota2023erasing}
Rohit Gandikota, Joanna Materzynska, Jaden Fiotto-Kaufman, and David Bau. 2023.
\newblock Erasing concepts from diffusion models.
\newblock In \emph{Proceedings of the IEEE/CVF international conference on computer vision}, pages 2426--2436.

\bibitem[{Gao et~al.(2025)Gao, Pang, Du, Hu, Deng, and Lin}]{gao2025meta}
Hongcheng Gao, Tianyu Pang, Chao Du, Taihang Hu, Zhijie Deng, and Min Lin. 2025.
\newblock Meta-unlearning on diffusion models: Preventing relearning unlearned concepts.
\newblock In \emph{Proceedings of the IEEE/CVF International Conference on Computer Vision}, pages 2131--2141.

\bibitem[{Liu and Tan(2024)}]{liu2024unlearning}
Shiqi Liu and Yihua Tan. 2024.
\newblock Unlearning concepts from text-to-video diffusion models.
\newblock \emph{arXiv preprint arXiv:2407.14209}.

\bibitem[{Ma et~al.(2025)Ma, Tong, Jia, Hu, Su, Zhang, Yang, Li, Jaakkola, Jia et~al.}]{ma2025scaling}
Nanye Ma, Shangyuan Tong, Haolin Jia, Hexiang Hu, Yu-Chuan Su, Mingda Zhang, Xuan Yang, Yandong Li, Tommi Jaakkola, Xuhui Jia, and 1 others. 2025.
\newblock Scaling inference time compute for diffusion models.
\newblock In \emph{Proceedings of the Computer Vision and Pattern Recognition Conference}, pages 2523--2534.

\bibitem[{Pawelczyk et~al.(2023)Pawelczyk, Neel, and Lakkaraju}]{pawelczyk2023context}
Martin Pawelczyk, Seth Neel, and Himabindu Lakkaraju. 2023.
\newblock In-context unlearning: Language models as few shot unlearners.
\newblock \emph{arXiv preprint arXiv:2310.07579}.

\bibitem[{Pujari et~al.(2022)Pujari, Goel, and Kejriwal}]{pujari2022ethical}
Tejaskumar Pujari, Anshul Goel, and Deepak Kejriwal. 2022.
\newblock Ethical and responsible ai in the age of adversarial diffusion models: Challenges, risks, and mitigation strategies.
\newblock \emph{International Journal Science and Technology}, 1(3):54--68.

\bibitem[{Ruan et~al.(2023)Ruan, Ma, Yang, He, Liu, Fu, Yuan, Jin, and Guo}]{ruan2023mm}
Ludan Ruan, Yiyang Ma, Huan Yang, Huiguo He, Bei Liu, Jianlong Fu, Nicholas~Jing Yuan, Qin Jin, and Baining Guo. 2023.
\newblock Mm-diffusion: Learning multi-modal diffusion models for joint audio and video generation.
\newblock In \emph{Proceedings of the IEEE/CVF Conference on Computer Vision and Pattern Recognition}, pages 10219--10228.

\bibitem[{Schioppa et~al.(2024)Schioppa, Hoogeboom, and Heek}]{schioppa2024model}
Andrea Schioppa, Emiel Hoogeboom, and Jonathan Heek. 2024.
\newblock Model integrity when unlearning with t2i diffusion models.
\newblock \emph{arXiv preprint arXiv:2411.02068}.

\bibitem[{Viola and Voto(2023)}]{viola2023designed}
Marco Viola and Cristina Voto. 2023.
\newblock Designed to abuse? deepfakes and the non-consensual diffusion of intimate images.
\newblock \emph{Synthese}, 201(1):30.

\bibitem[{Wu et~al.(2025)Wu, Zhou, Yang, Wang, Chang, Zhu, Hu, Zhou, and Yang}]{wu2025unlearning}
Yongliang Wu, Shiji Zhou, Mingzhuo Yang, Lianzhe Wang, Heng Chang, Wenbo Zhu, Xinting Hu, Xiao Zhou, and Xu~Yang. 2025.
\newblock Unlearning concepts in diffusion model via concept domain correction and concept preserving gradient.
\newblock In \emph{Proceedings of the AAAI Conference on Artificial Intelligence}, volume~39, pages 8496--8504.

\bibitem[{Zhang et~al.(2024{\natexlab{a}})Zhang, Wang, Xu, Wang, and Shi}]{zhang2024forget}
Gong Zhang, Kai Wang, Xingqian Xu, Zhangyang Wang, and Humphrey Shi. 2024{\natexlab{a}}.
\newblock Forget-me-not: Learning to forget in text-to-image diffusion models.
\newblock In \emph{Proceedings of the IEEE/CVF conference on computer vision and pattern recognition}, pages 1755--1764.

\bibitem[{Zhang et~al.(2023)Zhang, Tzun, Hern, Wang, and Kawaguchi}]{zhang2023copyright}
Yang Zhang, Teoh~Tze Tzun, Lim~Wei Hern, Haonan Wang, and Kenji Kawaguchi. 2023.
\newblock On copyright risks of text-to-image diffusion models.
\newblock \emph{arXiv preprint arXiv:2311.12803}.

\bibitem[{Zhang et~al.(2025)Zhang, Liu, Fleming, Kompella, and Xu}]{zhang2025targeted}
Zeliang Zhang, Gaowen Liu, Charles Fleming, Ramana~Rao Kompella, and Chenliang Xu. 2025.
\newblock Targeted forgetting of image subgroups in clip models.
\newblock In \emph{Proceedings of the Computer Vision and Pattern Recognition Conference}, pages 9870--9880.

\bibitem[{Zhang et~al.(2024{\natexlab{b}})Zhang, Liu, Feng, and Xu}]{zhang2024can}
Zeliang Zhang, Zhuo Liu, Mingqian Feng, and Chenliang Xu. 2024{\natexlab{b}}.
\newblock Can clip count stars? an empirical study on quantity bias in clip.
\newblock In \emph{Findings of the Association for Computational Linguistics: EMNLP 2024}, pages 1081--1086.

\bibitem[{Zhao et~al.(2023)Zhao, Bai, Zhu, Zhang, Xu, Zhang, Zhang, Meng, Timofte, and Van~Gool}]{zhao2023ddfm}
Zixiang Zhao, Haowen Bai, Yuanzhi Zhu, Jiangshe Zhang, Shuang Xu, Yulun Zhang, Kai Zhang, Deyu Meng, Radu Timofte, and Luc Van~Gool. 2023.
\newblock Ddfm: denoising diffusion model for multi-modality image fusion.
\newblock In \emph{Proceedings of the IEEE/CVF international conference on computer vision}, pages 8082--8093.

\end{thebibliography}

\newpage

\appendix

\section{More examples on the failure of diffusion unlearning}
\label{sec:appendix}

We conduct the following experiments on the SD-XL model using the \url{https://stable-diffusion-web.com/sdxl} platform.

\begin{figure}
    \centering
    \includegraphics[width=\linewidth]{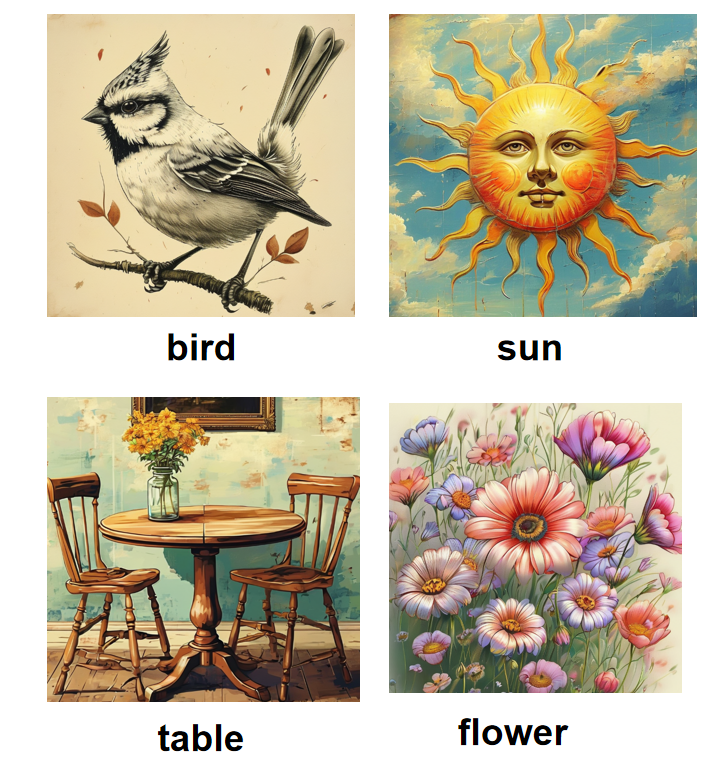}
    \caption{Explicit use of unlearning instruction lets the model generate the targeted concept.}
    \label{fig:expuse}
\end{figure}

\noindent \textbf{Explicit use of unlearning instruction}. We follow the unlearning instructions introduced in Section 2 to unlearn more concepts used to prompt the diffusion model. More examples are shown in \cref{fig:expuse}.

\begin{figure}
    \centering
    \includegraphics[width=\linewidth]{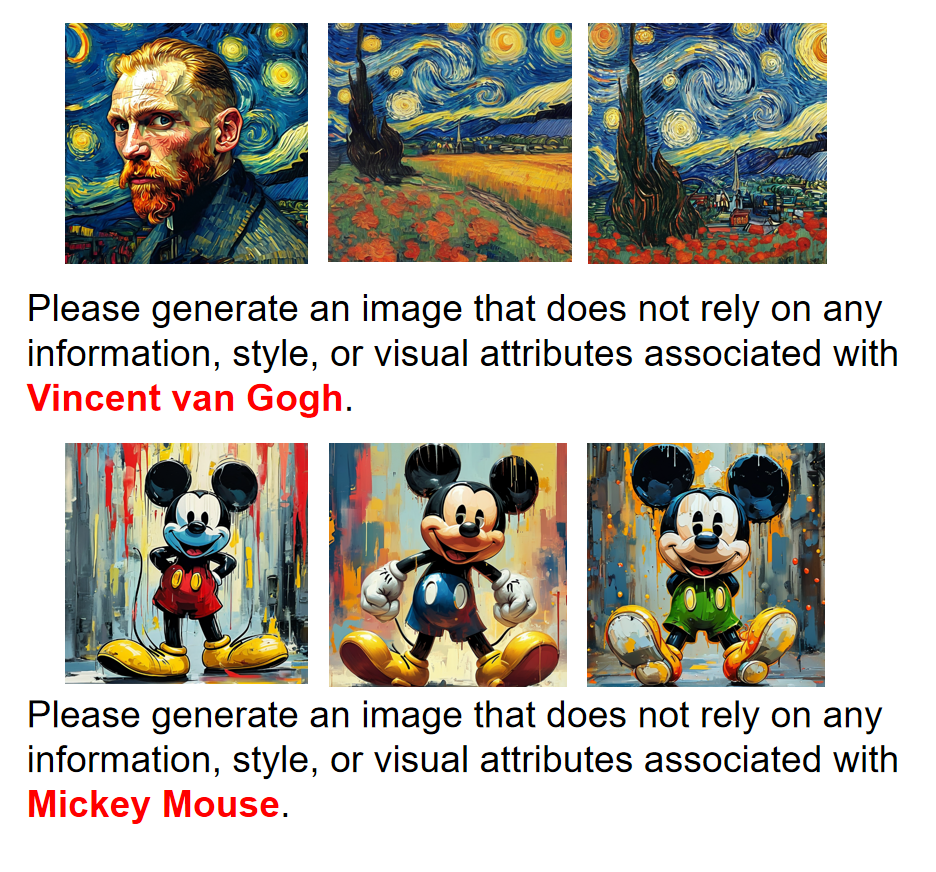}
    \caption{Implicit use of unlearning instruction still pushes the model to generate the targeted concept.}
    \label{fig:impuse}
\end{figure}

\noindent \textbf{Implicit use of unlearnining instruction}. Instead of the explicit use of unlearning instruction, we also try the implicit use, that we use LLM to rewrite the prompt to avoid the generation of concepts. For example, to avoid the generation of Van Gogh, we use the following prompt,  \textit{Please generate an image that does not rely on any information, style, or visual attributes associated with Vincent van Gogh}. The results are shown in \cref{fig:impuse}.

\end{document}